\documentclass{article}

\usepackage{microtype}
\usepackage{graphicx}
\usepackage{subfigure}
\usepackage{booktabs} % for professional tables

\usepackage{hyperref}
\usepackage{amsmath}
\usepackage{nicefrac}
\usepackage{array}
\usepackage[table]{xcolor}
\usepackage{hhline}
\usepackage{multirow}
\usepackage{caption}
\usepackage{tikz}
\usetikzlibrary{calc}
\usepackage[inline]{enumitem}

\usepackage{makecell}
\usepackage{rotating}
\usepackage{multirow}
\usepackage{amssymb}

\usepackage[accepted]{icml2020}

\icmltitlerunning{Learning to Simulate and Design for Structural Engineering}

\begin{document}

\twocolumn[
\icmltitle{Learning to Simulate and Design for Structural Engineering}
\begin{icmlauthorlist}
\icmlauthor{Kai-Hung Chang}{adsk}
\icmlauthor{Chin-Yi Cheng}{adsk}
\end{icmlauthorlist}
\icmlaffiliation{adsk}{Autodesk Research, San Francisco, California, United States}
\icmlcorrespondingauthor{Kai-Hung Chang}{kai-hung.chang@autodesk.com}
\icmlkeywords{Machine Learning, ICML, Graph Neural Network, Structured Prediction,}
\vskip 0.3in
]

\printAffiliationsAndNotice{}

\begin{abstract}
The structural design process for buildings is time-consuming and laborious. To automate this process, structural engineers combine optimization methods with simulation tools to find an optimal design with minimal building mass subject to building regulations. However, structural engineers in practice often avoid optimization and compromise on a suboptimal design for the majority of buildings, due to the large size of the design space, the iterative nature of the optimization methods, and the slow simulation tools. In this work, we formulate the building structures as graphs and create an end-to-end pipeline that can learn to propose the optimal cross-sections of columns and beams by training together with a pre-trained differentiable structural simulator. The performance of the proposed structural designs is comparable to the ones optimized by genetic algorithm (GA), with all the constraints satisfied. The optimal structural design with the reduced the building mass can not only lower the material cost, but also decrease the carbon footprint.

%The structural design process for buildings is time-consuming and laborious. To automate this process, structural engineers combine optimization methods with simulation tools to find an optimal design with minimal building mass. However, this process is computationally expensive and extremely slow because of the size of the design space, the iterative nature of the optimization methods, and the slow simulation tools. Therefore, structural engineers often avoid optimization or have to compromise on a suboptimal design for the majority of buildings. In this work, we present an end-to-end pipeline, intended to speed up the structural design process. We achieve this by formulating building structures as graphs and using the feedback from a differentiable simulator to train a graph neural network to propose the cross-sections of columns and beams. Our trained model, NeuralSizer, can directly output the design, and its performance is comparable to the genetic algorithm (GA), with all the constraints satisfied. The optimal structural design with the reduced the building mass can not only lower the material cost, but also decrease the carbon footprint.
\end{abstract}

\section{Introduction}

% a. Intro to the structural design process for buildings
%   i. simulation
%   ii. design
%   iii. expensive, not sustainable (can use material/numbers in Nicolas Mangon's talk)
% b. automating structural design
%   i. traditional optimization -> slow sim and opt
%   ii. we want to improve the speed for both sim and opt
% c. a proper representation for building structure: graph
% d. approach
%   show the diagram that represent our task: structural design optimization
% e. bridge the gap between opt and ML (TBD)
% f. boosting structural design process and make the design outcome environmental friendly
% g. release toy dataset for ML research on structural design

Structural design of buildings is to design the optimal structures subject to a design objective, such as minimizing material usage for cost and sustainability reasons. The design also has to satisfy a set of rules from established standards known as building codes, for example, a limited displacement under loading and seismic force. However, most structural engineers do not employ optimization in real-world cases for several reasons. First, the design space is large. A classic five-story building has typically over 500 columns and beams as design variables. Moreover, optimization algorithms usually require many iterations, and the evaluation for each optimization iteration takes 2 to 15 minutes to run structural simulation. As a result, a single optimization process can take days to converge, which does not fit the need for frequent design changes in a construction project. In practice, the majority of structural engineers complete structural designs with trial-and-error. After getting simulation results from a proposed design, structural engineers have to revise the design iteratively until all the building codes are satisfied. Since finding a valid design is already challenging and laborious, the outcome is usually over-designed, which means it satisfies the code and constraints but has poor performance in the design objective.

\begin{figure}[t!]
  \centering
  \includegraphics[width=\columnwidth]{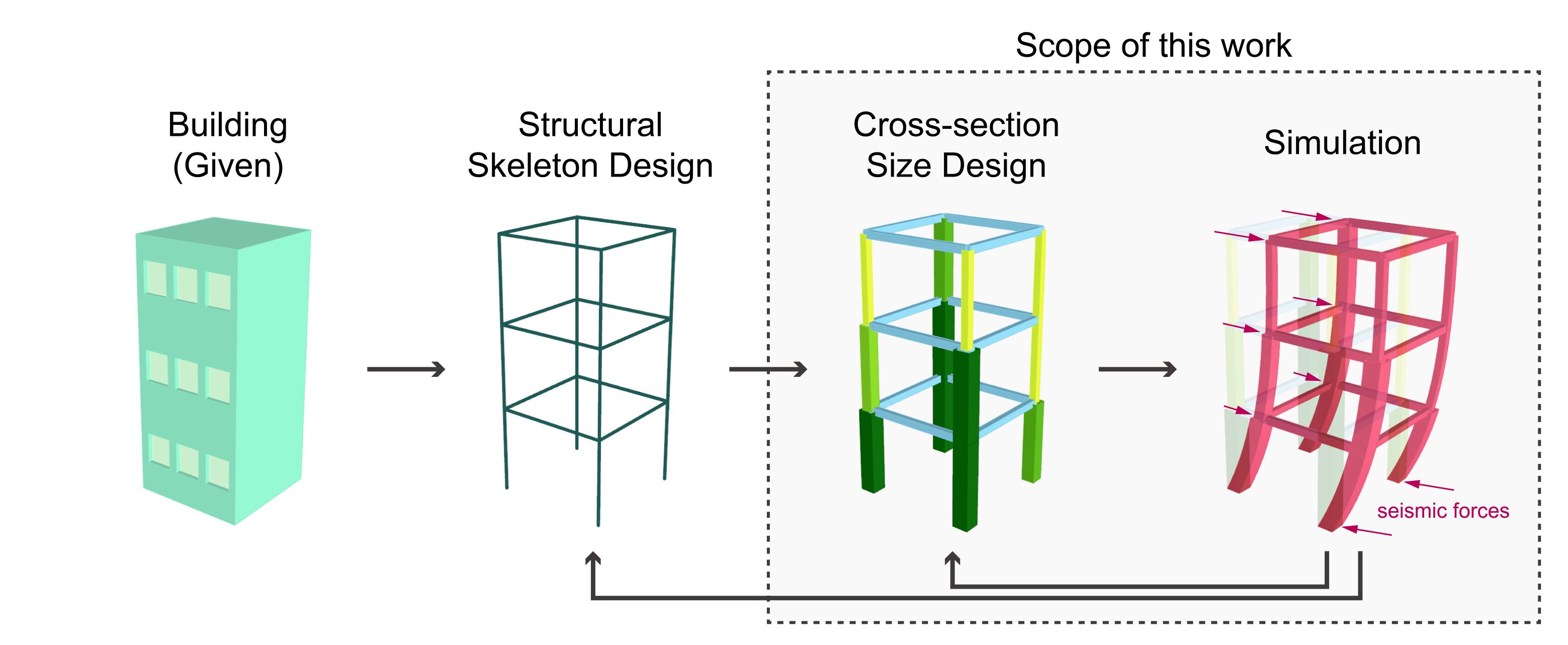}
  \caption{The iterative industrial structural design workflow.}
  \label{fig:workflow}
  \vskip -0.15in
\end{figure}
\begin{figure*}[h!]
  \centering
  \includegraphics[width=.8\textwidth]{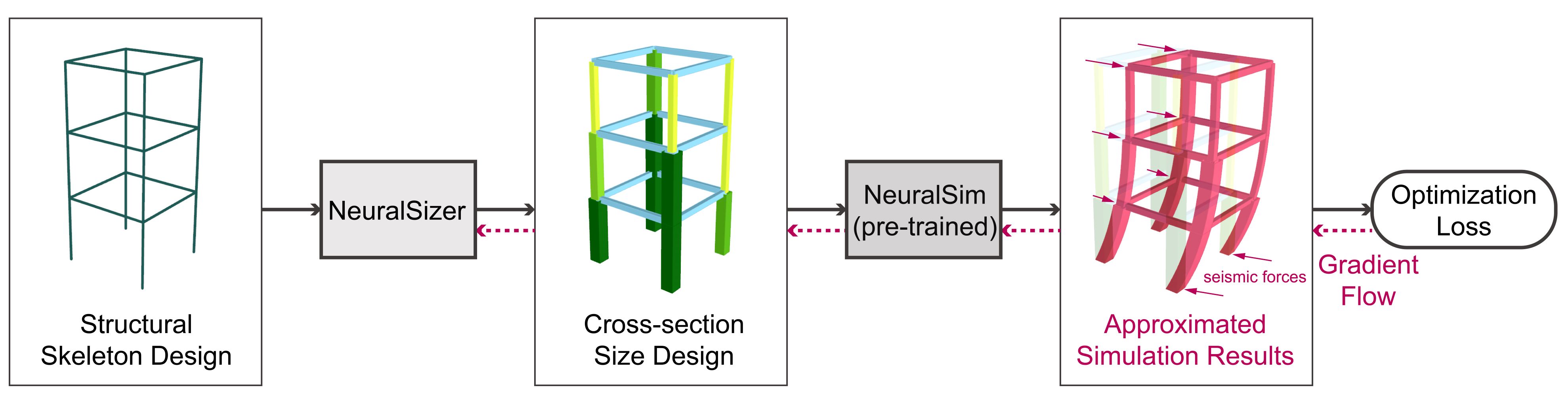}
  \caption{Our proposed end-to-end learning pipeline for solving the size design optimization problem.}
  \label{fig:pipeline}
\end{figure*}

The main contribution of this paper is proposing an end-to-end solution to automate the structural design process. As visualized in Figure \ref{fig:workflow}, a typical structural design process starts from a given building design, and then the structural engineer will propose a skeleton design, where the locations and connectivities of columns and beams are defined. After proposing the skeleton, the engineer will decide the cross-section size for each bar (column and beam). The engineer will then evaluate the design by running structural simulation and update the skeleton and cross-section sizes iteratively. As a starting point, the scope of this work only focuses on automating the size design as well as the structural simulation process. 

We discover that a building skeleton can be naturally represented as a graph and therefore propose two graph neural networks, NeuralSim and NeuralSizer. We train NeuralSizer to assign the optimal cross-section sizes to the given columns and beams and evaluate the size design using the pre-trained neural approximator, NeuralSim, instead of using real structural simulation tool. Taking advantage of the differentiable nature of NeuralSim, the gradient of the optimization loss can flow through it and update the learning parameters in NeuralSizer. The learning pipeline is illustrated in Figure \ref{fig:pipeline}. Our results show that NeuralSizer can produce design comparable to the design generated by genetic algorithm running for 1000 iterations. We also perform experiments including ablation, extrapolation, and user study.

\section{Related Work} \label{sec:pervious}
\subsection{Structural Optimization}
Most structural engineering research solves building design optimization problems with evolutionary algorithms, such as genetic algorithms \cite{rajeev1997genetic, balogh2012genetic, imran2019design}. These methods evaluate candidate solutions using a fitness function and iteratively evolve the solution population. The computational complexity is high due to the evaluation, especially with structural simulation tools. Latest studies that apply deep learning approaches use vector inputs to encode information of a structural component \cite{greco2018machine, torky2018deep, cheng2017optimal}, a single structure\cite{hasanccebi2013neural}, or 2D images to coarsely describe the structural geometry \cite{tamura2018machine}. These representations suffer from scalability and expressiveness and thus can hardly be applied to real-world cases.

\subsection{Graph Neural Network}
Graph neural networks (GNNs) have shown great successes in many learning tasks that require graph data and are applied to many new scientific domains including physics systems \cite{kipf2018neural, sanchez2018graph, watters2017visual}, fluid dynamics\cite{sanchez2020learning}, chemical molecules\cite{fout2017protein, jin2018learning, do2019graph}, and traffic networks \cite{guo2019attention, cui2019traffic}. For readers who are unfamiliar with GNNs, recent reviews and tutorials exist \cite{zhou2018graph, wu2020comprehensive}. Related to structural engineering domain, \cite{hamrick2018relational} uses reinforcement learning to train a graph neural network policy that learns to glue pairs of blocks to stabilize a block tower under gravity in a physics engine. However, the scale and the complexity of the tower (maximum of 12 blocks) are much smaller than real-world buildings.

Though some papers have applied GNNs to solve combinatorial optimization problems \cite{bello2016neural, kool2018attention, prates2019learning, li2018combinatorial}, applying GNN to solve design optimization problem is underexplored. For molecular design optimization problems, both \cite{jin2018junction} and \cite{you2018graph} train GNNs to generate a new molecular graph that optimizes objectives subject to constraints. \cite{jin2018junction} trains a junction tree variational autoencoder to obtain the latent embedding of a molecular graph and iteratively revise the latent embedding based on a neural prediction model. Constraints on the revision similarity is later enforced over a population of the revised molecular graphs to find the one that has the best predicted property. Similar to \cite{hamrick2018relational}, \cite{you2018graph} also chooses reinforcement learning approach and trains a graph convolutional policy network which outputs a sequence of actions to maximize the property reward. Compared to these two papers, our approach trains NeuralSizer to directly propose cross-sections, and uses NeuralSim as a differentiable simulator to provide a back-propagable loss.

\subsection{Differentiable Approximator for Design}
Differentiable approximators are commonly used to model the non-differentiable loss function of interest and to provide gradients in back-propagation. Taking drawing task as an example, StrokeNet \cite{zheng2018strokenet} and Canvas Drawer \cite{frans2018unsupervised} optimizes the unsupervised reconstruction loss with the aid of a differentiable renderer. The differentiable renderer models the relation between stroke actions and the output drawing. Given a target scene of3D shapes, \cite{tian2019learning} trains a neural program executer with a pre-trained scene generator, which maps the input code to the corresponding output scene. Outside of image generation domain, \cite{zhou2016model} takes a model-based approach by training a hand pose generation model followed by a non-linear layer, which acts as a differentiable approximator for forward kinematics computation.

% \subsection{ML as A Seed for Optimization (TBD)}
% d. (seeding for optimal/surrogate)

\section{Formulation}
% % ==============================================
\label{sec:formulation}

\subsection{Size Design Optimization Problem}
\label{sec:optimization_problem}
\begin{figure}[h!]
  \centering
  \includegraphics[width=.4\textwidth]{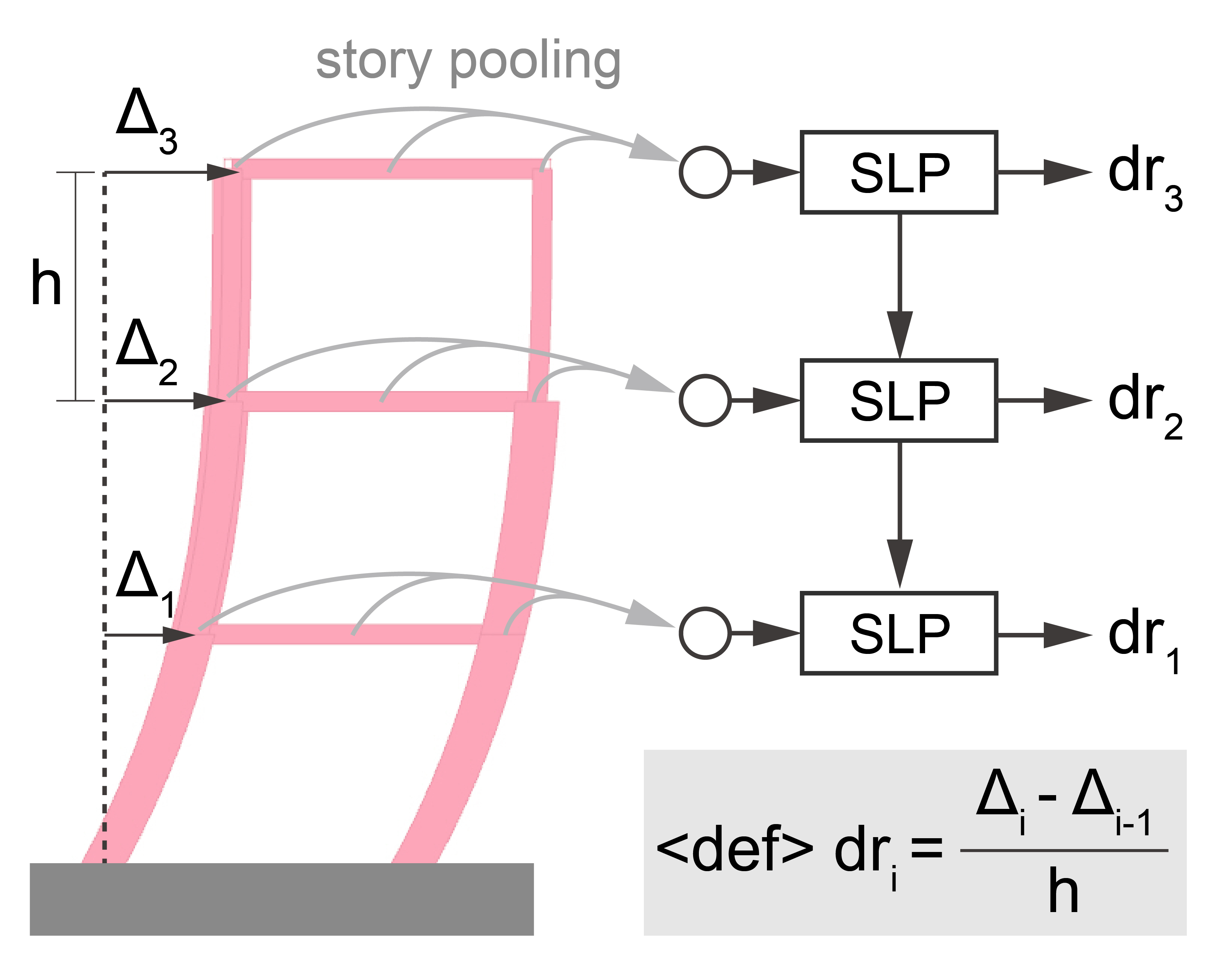}
  \caption{Drift Ratio is defined (in the grey text box) as the ratio of the relative lateral displacement of a story to the story height.}
  \label{fig:drift_ratio_def}
\end{figure}

In this paper, we target the size design optimization problem in the structural design process. After structure engineers complete the skeleton design, they have to decide the cross-sections for all the columns and beams, which directly impact the performance of the building, including the material usage, stability, constructability, etc. The size design optimization problem is formulated with the following objective and constraints:

\begin{itemize}
    \item Mass Objective: We want to minimize the total material mass of the building. 
    \item Drift Ratio Constraint: Building code regulations require a building to satisfy a set of constraints to ensure its stability and safety. The drift ratio constraint requires the drift ratios in all stories (visualized and defined in the grey text box in Figure \ref{fig:drift_ratio_def}) to be less than some limit under lateral seismic loads.
    \item Variety Constraint: This constraint comes from the constructability requirement which sets a maximum number for different cross-section types used. Using too many different cross-section types leads to higher manufacturing and transportation cost. 
\end{itemize}

In most cases, using stronger columns or beams improves stability, but leads to a larger total mass. The optimal design should satisfy both constraints and have a minimal building mass. Mathematical equations for the objective and constraints are provided in Section \ref{sec:sizer_loss}.

\subsection{Data Generation}
Due to the lack of real structural design data, we synthesize a dataset that contains building skeletons with randomly sampled cross-sections in real-world scale. We also use Autodesk Robot Structural Simulator (RSA), a simulation software widely used in the industry, to compute the structural simulation results for the synthetic dataset. Various loads are considered in the simulation: 
%To train NeuralSim as a differentiable simulator, we collect structural simulation data directly from Autodesk Robot Structural Analysis (RSA), a simulation software widely used in the industry. Due to a lack of real data, building geometries with random cross-sections are sampled in a construction site at a real-world scale. Various loads are considered:
\begin{enumerate*}
    \item Self-weight load of the building structure,
    \item Surface loads on floor panels which are distributed to the underlying beams,
    \item Surface loads on the roof story, and 
    \item Linear loads at the boundary beams for external walls.
\end{enumerate*}
NeuralSim is then trained with the building skeletons with paired cross-sections and simulated drift ratio. Please refer to the supplementary material for more details.

\subsection{Representing Building Structures as Graphs}

\begin{figure}[h!]
\begin{center}
\centerline{\includegraphics[width=\columnwidth]{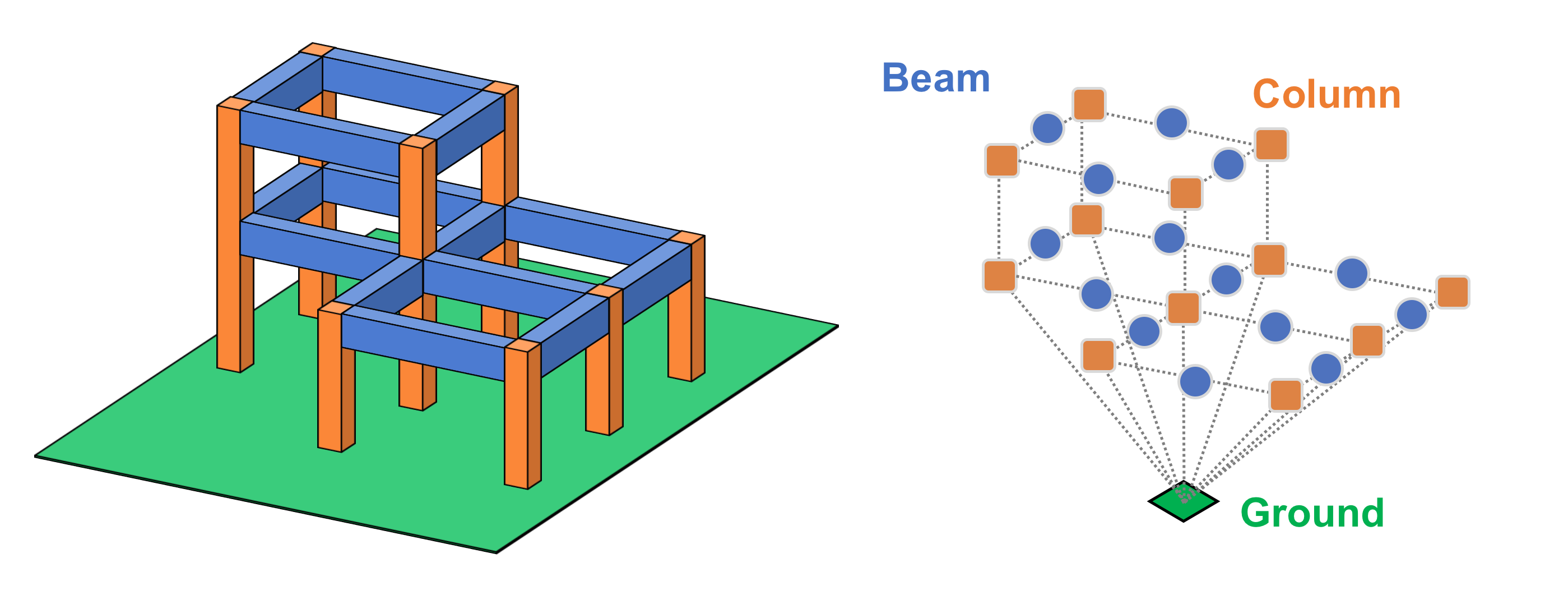}}
\caption{An example building structure and its structural graph representation. }
\label{fig:struct_graph}
\end{center}
\vskip -0.15in
\end{figure}

We represent building geometries as structural graphs. Every bar (column or beam) is represented as a graph node. An edge connects two nodes if the two corresponding bars are joined together. Information of bar $i$ is stored as node feature $v_{i}=[p_{1}, p_{2}, B, T, L]$, where $p_{1}$ and $p_{2}$ locates the two endpoints of the bar, $B$ indicates if the bar is a beam or a column, $T$ is a one-hot vector representing the cross-section, and $L$ provides auxiliary loading condition information, including 
\begin{enumerate*}
    \item if the bar is on the roof story,
    \item if the bar is on the boundary, and 
    \item the surrounding floor penal areas which are multiplied by the per-area loads when computing the surface loads.
\end{enumerate*}
A pseudo ground node is connected to all first-story columns and the values of its feature vector are all -1. The structural graph of a simple example structure is illustrated in Figure \ref{fig:struct_graph}. Story level indices of each bar are also saved. 
% Edge features and graph features are not used, but they can represent, for instance, structural connector properties and site information, respectively, in more complex use cases.

\section{Graph Neural Networks}
% % ==============================================
\label{sec:GNN}

\subsection{NeuralSim}
\label{sec:neuralsim}
\subsubsection{Network Structure}
Inspired by GraphNet \cite{battaglia2018relational}, NeuralSim also contains three steps: encoder, propagation, and decoder. The encoder first maps the input node features into the embedding space. Then, in the propagation step, each node embedding is iteratively updated based on the aggregated message, which is computed using the neighbor embeddings as well as the current node embedding. We also experiment a model variant called NeuralSim + PGNN, which extends the message by integrating the position-aware message \cite{you2019position}. Though the classic message function can model the forces and reaction forces between bars, position-aware message can further provide global information that helps identify loading conditions, such as if a bar is on the roof story or boundary. After a fixed number of propagation steps, we obtain the final embeddings of all nodes. Given the story indices, story embeddings are computed by pooling over all final embeddings of nodes in the same story. 

Instead of using a standard multi-layer perceptron (MLP) as a decoder, we design a \textit{Structured Decoder} (SD) for outputting drift ratios in each story. The illustration of SD is visualized in Figure \ref{fig:drift_ratio_def}. Starting from the roof story, SD updates each story embedding in the top-down order. The update module takes in the embedding of a story and the updated embedding of the above story to replace the current embedding. The design of SD is intended to mimic the algorithmic structure of the drift ratio's definition. Moreover, given the physics nature that the lower the story, the higher the story drift, drift ratio in the same story will have different distributions if two buildings have different total numbers of stories. After updating the story embeddings, we pass them to two MLP decoders. One decoder outputs the predicted drift ratios while the other classifies if the ground-truth drift ratios exceed the drift ratio limit 0.015 or not. 

\subsubsection{Loss}
Given the two kinds of output from the decoders, our multi-task loss is defined as the sum of an L1 loss for the drift ratio output and a binary cross-entropy loss for the classifier output. Later in the result section, we show that this multi-task loss helps NeuralSim learn a better embedding. 

\subsubsection{Training}
We split the total 4000 data into 3200, 400, and 400 for training, evaluation, and testing purposes. Adam Optimizer is used with learning rate 1e-4 and weight decay 5e-4. Batch size is set to 1 and the number of epoch is 5.

\subsection{NeuralSizer}
\subsubsection{Network Structure}
Similar network structure is adopted by NeuralSizer. It has the same encoder and propagation step. SD is not used since there isn't a strong bias in the size design. Instead, a graph embedding is computed to provide the information of the entire structure. In the end, an MLP decoder is used to map the final node embeddings concatenated by the graph embedding to the probability over cross-sections using hard Gumbel Softmax function \cite{jang2016categorical}. The hard version returns deterministic samples as one-hot vectors to ensure consistency of NeuralSim inputs, but in back-propagation, the derivatives are calculated as if they are the soft samples.

\subsubsection{loss}
Given the objective and constraints of the optimization problem in \ref{sec:optimization_problem}, we formulate them as differentiable losses. 

\begin{itemize}
    \item Mass Objective $obj$: The total mass of a bar is the product of its length, the area of its cross-section, and the material density. The length is derived from the two endpoint locations of the bar and we assume all bars are made of the same steel. The total mass is normalized by the number of bars in the structure.
    \item Drift Ratio Constraint $l_{dr}$: 
    This constraint requires the absolute value of all drift ratio $dr_{i}$ to be less than a limit $lim$. Therefore, we penalize the mean of how much the drift ratio in each story exceeds the limit:
    \[
    l_{dr} = Mean\{\text{LeakyReLU}(|dr_{i}|-lim)\} \leq 0
    \]
    \item Variety Constraint $l_{var}$:
    The variety constraint penalizes the number of cross-section usages more than 6. We compute the usage percentage of each cross-section $p$ and expect the sum of top 6 percentages to be 1. In other words, we can formulate the constraint as below:
    \[
    l_{var} = 1-\text{SumTop6}(p) = 0
    \]
    \item Entropy Constraint $l_{H}$:
    To avoid quick overfitting to some undesired local minimum, we introduce this entropy constraint inspired by maximal entropy reinforcement learning (RL) \cite{haarnoja2018soft}. Denote the entropy of NeuralSizer output of each bar as $H_{i}$, the maximum entropy over $n$ different cross-sections $H_{max}$, and a target ratio $\alpha=0.6$. The entropy constraint is formulated as:
    \[l_{H} = Mean\{H_{i}\}/H_{max} - \alpha = 0\]
    Without this entropy constraint, NeuralSizer converges within 50 iterations in the experiment and always uses one cross-section type for all bars.
\end{itemize}

The total loss $L$ equals $w_{0}obj + w_{1}l_{dr} + w_{2}l_{var}+w_{3}l_{H}$. Instead of fine-tuning the weights $w_{i}$ manually, we automate the process by optimizing the dual objective and approximating dual gradient descent\cite{boyd2004convex}. This technique has shown successful results in soft actor-critic algorithms \cite{haarnoja2018soft} and reward constrained policy optimization \cite{achiam2017constrained}. A brief explanation of the method is given in the supplementary material.

\label{sec:sizer_loss}
\subsubsection{Training}
In each epoch, a new structural graph is randomly generated and fed to NeuralSizer to get the design output. The output cross-sections are concatenated to the node embeddings in the structural graph, which is passed to NeuralSim for structural simulation. Given the design output and drift ratio output, the total loss is computed. NeuralSizer updates based on the back-propagation gradients once every 5 epochs, and runs 50,000 epochs for training. Though having the best accuracy, NeuralSim + PGNN has a much longer inference time than NeuralSim due to the computation of the position-aware message. Therefore, we use a frozen pre-trained NeuralSim, which also shows high accuracy throughout the training.

\section{Experiments} 
\label{sec:exp}
All training and testing run on a Quadro M6000 GPU.

\subsection{NeuralSim Results}
We compare NeuralSim to four other graph neural network models: GCN \cite{kipf2016semi}, GIN \cite{xu2018powerful}, GAT \cite{velivckovic2017graph}, and PGNN \cite{you2019position}. Table \ref{table:NeuralSimResults} shows the L1 loss and the relative accuracy of the drift ratio output as well as the classification accuracy of the classifier output. NeuralSim trained with the Structured Decoder (SD) outperforms GCN, GIN, GAT, and PGNN in all three metrics. Moreover, integrating the position-aware message from PGNN helps further improve the performance.

Ablation study results are also included in Table \ref{table:NeuralSimResults}. NeuralSim shows better performance when trained with SD since the imposed inductive bias of SD models the increasing drift ratios in lower stories. Moreover, adding the classifier output and the binary cross-entropy loss helps NeuralSim learn a better embedding and thus improves the performance.

A plot of all learning curves is included in the supplementary material. Training takes around 3.5 hours and a forward propagation of NeuralSim for one design takes 6.789 milliseconds in average. Compared to our simulation software Autodesk RSA, which takes 13 seconds, NeuralSim is 1900 times faster with 97.36\% accuracy. NeuralSim + PGNN takes 43.92 milliseconds, which is 300 times faster. 

\begin{table*}[h!]   %Qualitative table
\caption{NeuralSim Performance compared to Other Models}
\label{table:NeuralSimResults}
\centering
\begin{tabular}{l|ccc}
    \toprule
    Model & L1 Loss $\times 1\mathrm{e}{-4}$ & Relative Accuracy & Classification Accuracy\\
    \midrule
    GCN & 16.01 & 94.86 & 89.22 \\ % ID: 031102
    GIN & 33.85 & 89.62 & 84.27 \\ % ID: 130827
    GAT & 10.87 & 96.41 & 93.35 \\ % ID: 220202
    PGNN & 9.39 & 96.72 &  94.83 \\%  ID: 161448 
    NeuralSim &  \textbf{7.57} & \textbf{97.36} & \textbf{95.64} \\ % ID: 041152
    NeuralSim + PGNN & \textbf{5.01} & \textbf{98.22} & \textbf{96.43} \\%  ID: 161440 
    \midrule
    NeuralSim(no SD) & 10.24  & 96.65 & 92.71 \\%       ID: 104338
    NeuralSim(only L1 loss) & 16.47 & 95.24 &  n/a \\%  ID: 155228 
    \bottomrule
\end{tabular}
\end{table*}

In Table \ref{table:ExtrapolateNeuralSim}, NeuralSim demonstrates its generalizability to test data beyond the training scope. We split the test data into 3 buckets based on the numbers of stories: 1$\sim$3, 4$\sim$7, and 8$\sim$10 story. One model is trained with 1$\sim$10 story buildings while the other is trained with 4$\sim$7 story buildings. Both models are trained with the same amount of training data and tested against each bucket. The results in the first row show that the performance variation across different buckets is small. The second row demonstrates that NeuralSim also performs well on extrapolated data. The learned message passing module in NeuralSim models the physics of force propagation, which is universal across buildings of different numbers of stories. As a result, NeuralSim shows strong generalizability to extrapolated data. 

\begin{table*}[h!]
\caption{NeuralSim Generalizability Results}
\label{table:ExtrapolateNeuralSim}
\centering
\begin{tabular}{c|c|ccc}
    \toprule  
    Train Data & Test Data & L1 Loss $\times 1\mathrm{e}{-4}$ & Relative Accuracy & Classification Accuracy\\   % Test on 1~10 dataset [3600:]
    \midrule
    \multirow{3}{*}{\shortstack{1$\sim$10 story\\(Baseline)}} % 041152 
    &1$\sim$3 story   & 6.09& 98.09& 95.01\\
    &4$\sim$7 story   & 6.67& 97.55& 96.31\\
    &8$\sim$10 story  & 8.73& 96.34& 96.79\\
    \hline
    \multirow{3}{*}{\shortstack{4$\sim$7 story}} % 014820
    &1$\sim$3 story   & 23.97 & 91.70& 85.75\\
    &4$\sim$7 story   & 15.49& 92.93& 96.40\\
    &8$\sim$10 story  & 26.43& 91.11& 84.33\\
    \bottomrule
\end{tabular}
\end{table*}

\subsection{NeuralSizer Results}
Two scenarios are created. The high safety factor scenario has a drift ratio limit 0.015 while the low safety factor scenario has a drift ratio limit 0.025. For each scenario, two experiments are conducted with different weights on the mass objective. Table \ref{table:NeuralSizerResults} summarizes the results of all experiments. All constraints are close to zero, indicating that NeuralSizer learns to satisfy the hard constraints. In particular, the drift ratio constraint, which measures how much the drift ratio exceeds the limit, is negligible compared to the magnitude of drift ratios. Several designs are visualized in Figure \ref{fig:NeuralSizerViz}. Training takes around 2.5 hours and a forward propagation of NeuralSizer for one design takes 10.07 milliseconds in average.

\begin{table*}[h!]  
\caption{NeuralSizer Results under Different Scenarios}
\label{table:NeuralSizerResults}
\centering
\begin{tabular}{|c|c|c|c|c|}
    \hline
    \multirow{2}{*}{Scenario} & \multirow{2}{*}{Objective Weight}
    & Objective & \multicolumn{2}{c|}{Constraints}\\\cline{3-5} 
    & &Mass Objective  & Drift Ratio Constraint & Variety Constraint\\
    \hline
    \multirow{2}{*}{High Safety Factor} 
    % &0.01   &  & $\times 1\mathrm{e}{-7}$ &  $\times 1\mathrm{e}{-8}$ \\   % 
    &1  & 0.870 & 6.00$\times 1\mathrm{e}{-7}$ & 0.01$\times 1\mathrm{e}{-8}$ \\  % 111510
    &10 & 0.735 & 1.34$\times 1\mathrm{e}{-7}$ & 1.04$\times 1\mathrm{e}{-8}$ \\  % 175657
    \hline
    \multirow{2}{*}{Low Safety Factor} 
    % &0.01   & $\times 1\mathrm{e}{-7}$ &  $\times 1\mathrm{e}{-8}$ \\   % 
    &1  & 0.592 & 6.42$\times 1\mathrm{e}{-5}$ &  1.67$\times 1\mathrm{e}{-8}$ \\   % 213211
    &10 & 0.596 & 3.32$\times 1\mathrm{e}{-5}$ &  1.78$\times 1\mathrm{e}{-8}$ \\   % 161804
    \hline
\end{tabular}
\end{table*}

To test the generalizability of NeuralSizer, we pick the high safety factor scenario and objective weight = 10 for an experiment. Again, we compare two models (training with 1$\sim$10 story and 4$\sim$7 story buildings) and measures the performance of three test buckets (1$\sim$3, 4$\sim$7, and 8$\sim$10 story buildings). The results are summarized in Table \ref{table:ExtrapolateNeuralSizer}. Both models satisfy the constraints well and show similar performances across the three different buckets. NueralSizer trained with 4$\sim$7 story buildings also shows generalizability to buildings with more and less numbers of stories than the training range. 
\begin{table*}[h!]
\caption{NeuralSizer Generalizability Results (High Safety Factor, Objective Weight = 10)}
\label{table:ExtrapolateNeuralSizer}
\centering
\begin{tabular}{|c|c|c|c|c|}
    \hline
    \multirow{2}{*}{Train Data} & \multirow{2}{*}{Test Data}
    & Objective & \multicolumn{2}{c|}{Constraints}\\ \cline{3-5} 
    & &Mass Objective  & Drift Ratio Constraint & Variety Constraint\\
    \hline
    \multirow{3}{*}{\shortstack{1$\sim$10 story\\(Baseline)}}
    % 175657 
    &1$\sim$3 story   & 0.738 & 1.62$\times 1\mathrm{e}{-7}$ & 0.80$\times 1\mathrm{e}{-8}$ \\
    &4$\sim$7 story   & 0.725 & 1.28$\times 1\mathrm{e}{-7}$ & 0.97$\times 1\mathrm{e}{-8}$ \\
    &8$\sim$10 story  & 0.711 & 1.69$\times 1\mathrm{e}{-7}$ & 1.06$\times 1\mathrm{e}{-8}$ \\
    \hline
    \multirow{3}{*}{\shortstack{4$\sim$7 story}} 
    %  165628
    &1$\sim$3 story   & 0.773 & 2.96$\times 1\mathrm{e}{-7}$  & 1.30$\times 1\mathrm{e}{-8}$\\
    &4$\sim$7 story   & 0.746 & 3.50$\times 1\mathrm{e}{-7}$  & 1.25$\times 1\mathrm{e}{-8}$\\
    &8$\sim$10 story  & 0.728 & 3.68$\times 1\mathrm{e}{-7}$  & 1.01$\times 1\mathrm{e}{-8}$\\
    \hline
% amount of data distribution [376. 406. 387. 380. 422. 433. 417. 383. 382. 414.]
\end{tabular}
\end{table*}

% \subsection{Failure Cases}
% PGNN?

\subsection{User Feedback and User Study}
\label{sec:NeuralSizerViz}
\begin{figure*}[h!]
  \centering
  \includegraphics[width=.8\textwidth]{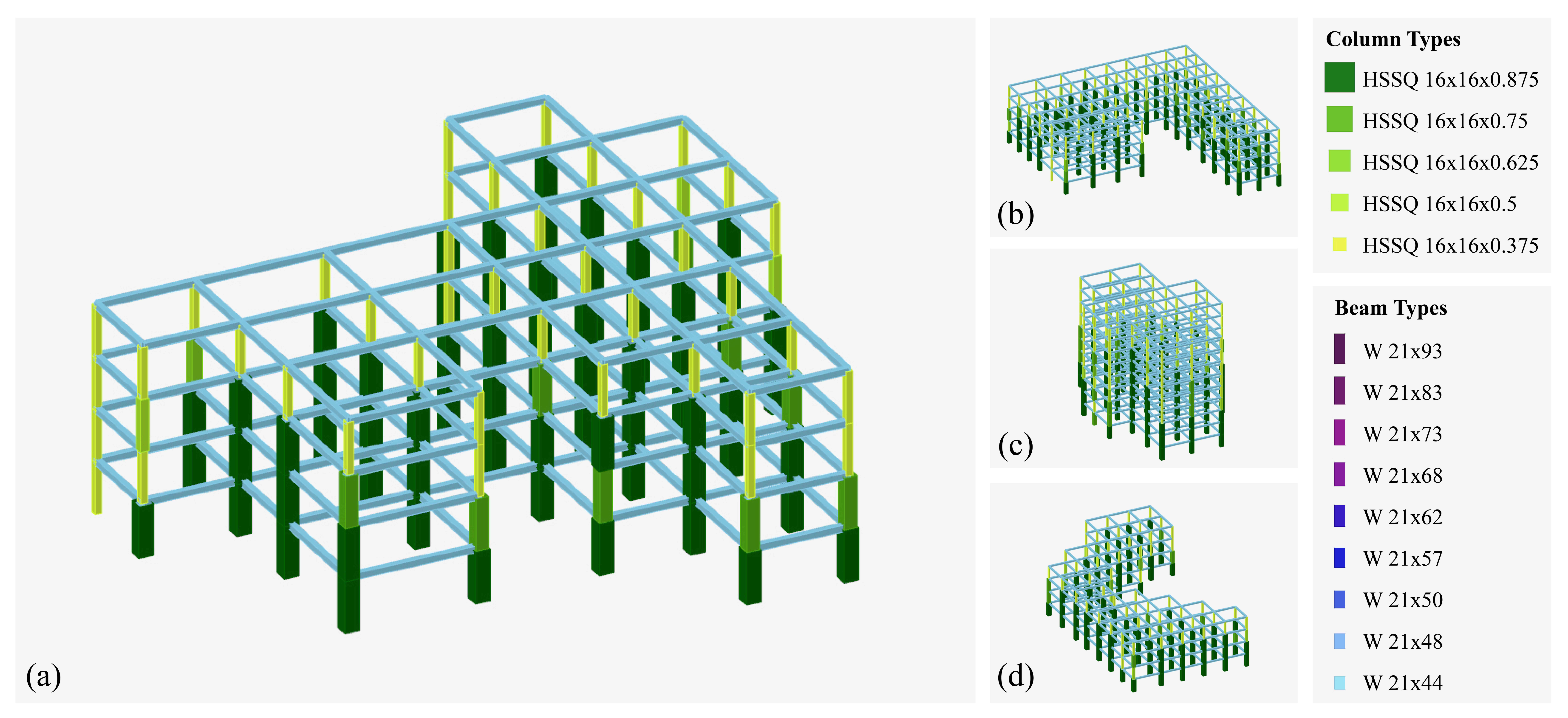}
  \caption{Visualization of NeuralSizer's Design Output}
  \label{fig:NeuralSizerViz}
\end{figure*}
Figure \ref{fig:NeuralSizerViz} visualizes the design outputs of NeuralSizer (high safety factor + objective weight 10) for various buildings. A cross-section with stronger structural properties is visualized in a darker color and a thicker stick.

To understand the quality from a professional perspective, we show 10 different designs to structural engineers and ask for their feedback. They say that the cross-section choices look natural to them, except that the design is too sophisticated as they usually assign same cross-sections in groups. Interestingly, structural engineers reveal some of the design rules that NeuralSizer reasons and learns and that they think are reasonable. These rules are listed and explained below.

\begin{itemize}
    \item During an earthquake, columns must support vertical gravity loads while undergoing large lateral displacements. Therefore, to satisfy the drift ratio constraint, NeuralSizer learns to distribute masses on columns more than beams.
    \item Since gravity loads are carried down the building structure, the loads accumulate and increase on lower stories. Columns for lower stories have a higher strength requirement than for higher stories. This can also be observed from NeuralSizer's design outputs.
    \item It is reasonable to have similar patterns in design outputs of different buildings. Given the objective and constraints in the optimization problem, structural engineers will probably design with similar patterns.
\end{itemize}

One user testing is conducted to compare our results with human design process. Given a building and 30 minutes, a structural engineer tries five design iterations. The first three iterations are able to satisfy all the constraints while the fourth and the fifth cannot. The best out of the three valid designs is used to compare with NeuralSizer results. The result shows NeuralSizer's design output has a better performance. The quantitative user testing results are provided in the supplementary material. The goal of this testing is not to show NeuralSizer can replace professional structural engineers, but to show that NeuralSizer can speed up the iterative design process by providing a better starting point.

\subsection{Comparison with Genetic Algorithm(GA)}
\subsubsection{GA Setup}
In this subsection, we compare our method with the genetic algorithm (GA), which is a widely-used algorithm in the structural optimization research. GA contains six steps:

\begin{enumerate}
    \item Initialization: 
    To start GA, a population of 100 candidate solutions is generated. A candidate has a chromosome which encodes the cross-sections in genes.
    \item Evaluation: 
    We compute the score of each candidate as measured by the total loss equation. The same initial weights for individual losses are used as when training NeuralSizer, except they are now fixed.
    \item Selection:
    The top five candidates are directly passed to the next generation without crossover and mutation. 
    \item Crossover: 
    We use a selection mechanism that random samples two candidates and outputs the better one, called selected candidate. With a crossover rate of 0.9, this selected candidate pairs with another selected candidate to
    breed a children candidate using crossover operation; otherwise, the selected candidate becomes the children candidate. In total, ninety-five children candidates are generated for the next population.
    \item Mutation:
    The genes in the ninety-five children candidates have a probability of 0.01 to randomly change to a different cross-section. After the mutation process, the ninety-five children candidates and the top five candidates form a new population for the next generation.
    \item Termination:
    The iteration process is terminated after a certain number of iterations or a limited amount of time. 
\end{enumerate}

\begin{table}[t!]
    \caption{Time Comparison of GA under Different Setups}
    \label{table:GAtime}
    \begin{tabular}{p{2.4cm}p{1.4cm}p{1.2cm}p{1.7cm}}
    \toprule
    Setup & Time & Total \newline Iterations & Time / \newline Iteration\\
    \midrule
    NeuroSizer & 10.07 ms & - & -\\
    GA + RSA   & 24 hr & 30 & -\\    
    $\rightarrow (estimated)$   & 2 weeks & 1000 & 20.16 mins\\
    GA + NeuralSim & 30 mins & 1000 & 0.03 mins\\
    \bottomrule
  \end{tabular}
\end{table}

\subsubsection{GA Time Comparison}
In the evaluation step, computing scores requires running simulation tools, which is time-consuming. However, by replacing simulation tools with NeuralSim, we can reduce the running time of GA. Table \ref{table:GAtime} lists running times of GA solving a building with 622 bars under different setups. NeuralSizer outputs a design within 10.07 milliseconds. If we evaluate GA candidates with Autodesk RSA (GA + RSA), we can run 30 iterations in 24 hours. By estimation, it will take up to 2 weeks to finish 1000 iterations. However, if NeuralSim is used for evaluation(GA + NeuralSim), it only takes around 30 minutes to complete 1000 iterations. 

% As a result, running GA with simulation tools is slow and impractical. More importantly NeuralSim can help speed up

\subsubsection{Using NeuralSizer Outputs as GA Seeds}
We first run GA + NeuralSim with random seeds for 1000 iterations as a baseline. Due to the stochastic nature of GA, We execute 10 runs and save the result that has the minimal total loss. Next, we use the NeuralSizer output as GA seeds. Since NeuralSizer outputs the probability of cross-sections, there are two seeding strategies: best seed and sampled seeds. Best seed finds the best design output based on the highest probability and populates it to the initial population. On the other hand, sampled seeds sample from the probability to generate different seeds as the initial population. We also run GA + RSA runs for 24 hours each with random seeds and sampled seeds. 

\begin{figure*}[h!]
  \caption{Performance Curves of GA using Different Seeding Approaches}
  \label{fig:seeding_compare}
  \centering
  \includegraphics[width=.8\textwidth]{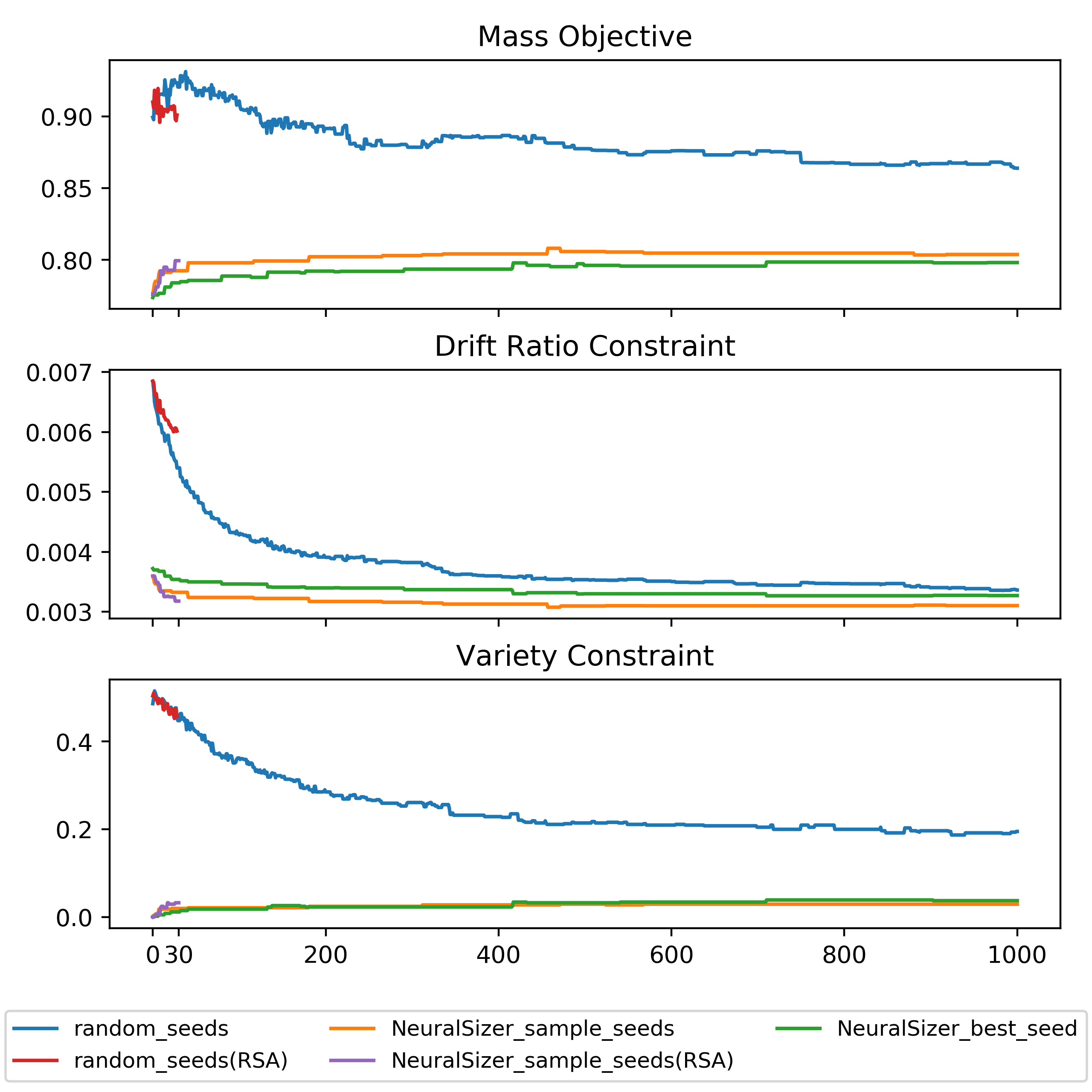}
\end{figure*}

Figure \ref{fig:seeding_compare} plots the performance curves of GA using different seeding approaches. All curves of GA using NeuralSizer seeds (orange, green, and purple curves) show lower losses at the beginning of GA. Moreover, after running the same number of iterations, curves starting with NeuralSizer seeds end up in better solutions than those starting with random seeds. However, the difference between using best seed and sampled seeds is little.

To quantitatively measure the performance across different designs, we run GA with NeuralSizer using sampled seeds and random seeds (once for each) over 20 different designs. Three metrics are defined and illustrated in Figure \ref{fig:metric}. 

For GA using random seeds, we denote the starting and ending losses as $(rand\textunderscore start,$ $ rand\textunderscore end)$. Likewise, for GA using NeuralSizer sampled seeds, they are denoted as $(sizer\textunderscore start, sizer\textunderscore end)$.

\begin{enumerate}
    \item $M_{1}=\dfrac{(rand\textunderscore start-sizer\textunderscore start)}{(rand\textunderscore start-rand\textunderscore end)}$\\
    The metric measures the percentage of improvement gained by initializing GA with NeuralSizer sampled seeds. If $M_{1}$ is larger than 100\%, it means that the best performance of the NeuralSizer sampled seeds at iteration 0 beats the best performance of GA with random seeds at iteration 1000.
    \item $M_{2}=\dfrac{(rand\textunderscore end - sizer\textunderscore end)}{rand\textunderscore end}$ \\
    The metric compared the best performances between GA with NeuralSizer sampled seeds and random seeds, both at 1000 iterations. A positive $M_{2}$ indicates that the final performance of GA with NeuralSizer sampled seeds is better.
    \item $M_{3}$ is the first iteration when $sizer\textunderscore start$ is less than $rand\textunderscore end$. Note that $M_{3}=0$ whenever $M_{1}>100\%$.
\end{enumerate}

\begin{table*}[h!]
\centering
\begin{minipage}{0.5\textwidth}
    \caption{NeuralSizer Seeding Performance}
    \label{tab:SeedingMetric}
	\centering
	\begin{tabular}{p{1.5cm}p{1.5cm}p{1.5cm}p{1.5cm}}
    \toprule
    Metric  & Mass  \newline Objective  & Drift Ratio  \newline Constraint & Variety \newline Constraint\\
    \midrule
    \multicolumn{2}{l}{High Safety Factor} & & \\
    1  & 232.60\% & 115.30\%  & 186.20\% \\   % drift limit 0.015
    % 2  & 8.02\%  & 21.45\%  & 100.00\% \\  
    2  &  7.43\%  & 25.70\%  & 95.82\% \\  
    3  & 0 & 25.6  & 0 \\  
    \midrule
    \multicolumn{2}{l}{Low Safety Factor} & & \\
    1  & 83.15\% & 95.35\%  & 156.22\% \\  
    % 2  & -4.80\%  & 99.76\%  & 99.99\% \\  
    2  &  4.16\%  & 49.22\%  & 32.53\% \\  
    3  & 128  & 0  & 0 \\  
    % 1  & 69.60\%  & 82.41\%  & 153.61\% \\  
    % 2  & -7.59\%  & 74.98\%  & 99.99\% \\  
    % 3  &  3.70\%  & 62.64\%  & 33.29\% \\  
    % 4  &  110 & 0 & 0 \\  % mention 110 iteration takes only 2 mins
    \bottomrule
  \end{tabular}
\end{minipage}\hfill
\begin{minipage}{0.45\textwidth}
    \captionof{figure}{Performance Metrics Illustration}
	\label{fig:metric}
    \centering
        \begin{tikzpicture}
        \node [above right,inner sep=0] (image) at (0,0) {\includegraphics[width=\textwidth]{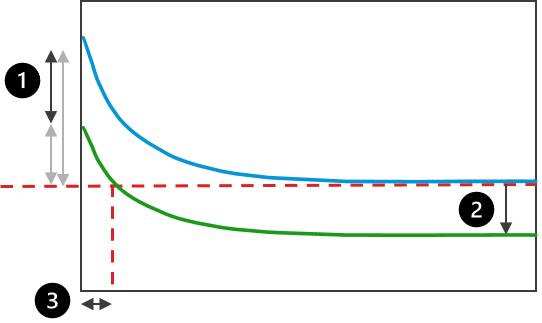}};
        % Create scope with normalized axes
        \begin{scope}[x={($0.1*(image.south east)$)},y={($0.1*(image.north west)$)}]
        % Grid
        % \draw[lightgray,step=1] (image.south west) grid (image.north east);
        \node [anchor=east, inner sep=0, text width =6.5cm, align=right] at (9.5,8.5) {This is just an illustration. \\Not a real curve!};
        \end{scope}
     \end{tikzpicture}
    
\end{minipage}
\end{table*}

The quantitative results are summarized in Table \ref{tab:SeedingMetric}. From $M_{1}$ results, we can see that some performance of NeuralSizer seeds are almost the same as the best performance of GA with random seeds at iteration 1000, and some are even better. This proves that NeuralSizer solves the size design optimization problem and that the results are comparable to GA. $M_{2}$ results show the capability of further optimizing the NeuralSizer output to obtain solutions which are better than optimizing with random seed after the same amount of iteration. Lastly, the maximum of the six $M_{3}$ values in Table \ref{tab:SeedingMetric} is 128. Thus, compared to the total 1000 iterations, using NeuralSizer seeds provide a speedup by at least 8 times.

\section{Conclusion}
\label{sec:conclusion}
In this paper, we propose an end-to-end learning pipeline to solve the size design optimization problem. Trained as a neural approximator for structural simulation, NeuralSim shows $\sim$ 97\% accuracy and is 1900 times faster than simulation tools. As a consequence, it can not only provide gradients to upstream models due to the differentiable nature, but also be used to provide quick evaluation in evolutionary solvers or instant feedback after every human design decision. NeuralSizer is trained to output optimal cross-sections subject to the optimization objective and constraints. The design outputs of NeuralSizer satisfy the constraint and are comparable to or sometimes better than the optimal solution of the genetic algorithm. Moreover, they can be used as initial seeds for the genetic algorithm, which speeds up the convergence and further optimizes the design. Last but not least, the design outputs demonstrate reasonable design rules and thus can also be used as the initial design for structural engineers to save design iterations.

The limitations and future work can be defined in three categories: first, the skeleton design phase in Figure \ref{fig:workflow} can be included into this pipeline and defined as a graph generation problem. Second, except from columns and beams, our graph representation does not cover all the structural components, such as walls and panels. Handling the heterogeneous attributes across the components is worth investigating.  Lastly, although the general structural constraints can be applied to any buildings in the world, the local building codes are varied. Converting the building code as part of the input is a potential solve. 

Buildings account for 40 percent of the global carbon dioxide emission \cite{/content/publication/world_energy_bal-2017-en}. Minimizing the mass of a building, we can not only reduce the material cost, but also decrease the carbon dioxide emission during the fabrication, transportation, and construction process, and therefore, make a huge impact to the environment. However, this domain is under-explored in the machine learning community. Through this work, we hope to arouse more research interest to explore this valuable direction. 

\section*{Awknowledgement}
We thank all reviewers who gave useful feedback and advice. Also, we would like to express appreciation to Dr. Mehdi Nourbakhsh who introduced this industry pain point and helped formulate the problem. Finally, special thanks to Waldemar Okapa and Grzegorz Skiba for their support in consulting about Autodesk Robot Structural Analysis and validating the data collection process.

\bibliography{paper_ref}
\bibliographystyle{icml2020}

\clearpage

\section*{Appendix}

\section{Data Collection}
This section describes the data collection process in detail. All unit abbreviations are listed in Table \ref{tab:unit}. We adopt the same beam spans, materials, cross-sections, and load cases used by a structural design company.

\subsection{Skeleton Creation}
Building skeleton are created by a fixed sampling algorithm due to the deficiency of real-world data. Each building is erected on a rectangular base which edges are sampled between 60 ft to 400 ft. A grid is created on the base and the intervals are sampled from the set of beam spans, ranging from 28 ft to 40 ft. A connected layout is sampled from the grid using depth-first-search algorithm which expands to neighboring grid cells with 0.5 probability. The same layout is vertically stacked up to 10 stories to form a voxel-like building geometry. Each voxel contains four columns on four vertical sides and four beams which form a rectangle frame on the top to support the floor panel. The story height is fixed at 16 ft.

\subsection{Structural Simulation Model in RSA}
Given the geometry of the building structure, we can create the corresponding structural simulation model in Autodesk Robot Structural Analysis (\href{https://www.autodesk.com/products/robot-structural-analysis/overview}{RSA}), which is an industrial structural simulation software. All the columns on the first story are not pinned, but fixed to the ground. Materials for columns and beams are Steel A500-46 and Steel A992-50 respectively. For each column and beam, the cross-section is randomly assigned from Table \ref{tab:cross-section}.  150 pcf Concrete floor panels are modeled as slabs on trapezoid plates with other parameters given in Table \ref{tab:panel}. The definition of the symbols can be found in this \href{https://knowledge.autodesk.com/support/robot-structural-analysis-products/learn-explore/caas/CloudHelp/cloudhelp/2018/ENU/RSAPRO-UsersGuide/files/GUID-2000B438-5453-4C90-B658-E3DE6F8AF33A-htm.html}{link}. In the graph representation, we do not model joists (smaller beams arranged in parallel across two beams to support floor panels). Instead, each surface load is converted to concentrated loads at joist endpoints. For each floor panel, three joists are placed across the longer edges.

% Unit Abbreviation
\begin{table}[b]  
\caption{Unit Abbreviation}
\label{tab:unit}
% \vskip 0.in
\begin{center}
\begin{small}  % \begin{sc}
\begin{tabular}{ll}
\toprule
Abbreviation & Full Unit  \\
\midrule
ft & Foot\\
pcf & Pound per cubic foot\\
psf & Pound per square foot\\
\bottomrule
\end{tabular}  % \end{sc}
\end{small}
\end{center}
\vskip -0.1in
\end{table}

% Cross-Section Library
\begin{table}[t] 
\caption{Cross-Section Library}
\label{tab:cross-section}
% \vskip 0.in
\begin{center}
\begin{small}  % \begin{sc}
\begin{tabular}{ll}
\toprule
Column & Beam  \\
\midrule
HSSQ 16x16x0.375 & W 21 x 44\\
HSSQ 16x16x0.5   & W 21 x 48\\
HSSQ 16x16x0.625 & W 21 x 50\\
HSSQ 16x16x0.75  & W 21 x 57\\
HSSQ 16x16x0.875 & W 21 x 62\\
& W 21 x 68\\
& W 21 x 73\\
& W 21 x 83\\
& W 21 x 93\\
\bottomrule
\end{tabular}  % \end{sc}
\end{small}
\end{center}
\vskip -0.1in
\end{table}

% Floor Panel Specification
\begin{table}[t!]   
\caption{Floor Panel Specification}
\label{tab:panel}
\vskip 0.in
\begin{center}
\begin{small}  % \begin{sc}
\begin{tabular}{ll}
\toprule
Parameter Name & Value\\
\midrule
        h & 6.3 in\\
        h1 & 2.56 in\\
        a & 7.4 in\\
        a1 & 1.73 in\\
        a2 & 4.96 in\\
        Th & 7.46 in\\
        Th 1 & 8.86 in\\
        Th 2 & 6.3 in\\
        \hline
        Joist Direction & Parallel to the shorter edge  \\
        \hline
        Material & Concrete\\
        Material Resistance & 3.5 ksi\\
        Material Unit Weight & 0.15 kip/ft3\\
        \hline
        Diaphragm & Rigid\\
        Load Transfer & Simplified one way\\
        Finite Element & None\\
\bottomrule
\end{tabular}  % \end{sc}
\end{small}
\end{center}
\vskip -0.2in
\end{table}

\subsection{Load Cases Setup}
IBC 2000 is the building code used for structural simulation. Below lists all the load cases:

\begin{enumerate}
    \item Self-Weight: This is the self-weight load acting in the gravitational direction for all structure elements. The coefficient is set to 1.1. \label{i-self}
    \item Super-Imposed Dead Load: Super-imposed dead load accounts for the static weight of the non-structure elements. Here we add 24 psf surface loads to all floor panels except the roof.\label{i-dead1}
    \item Live Load: Live load refers to the load that may change over time, such as people walking around. We consider 100 psf surface loasd on all floor panels except the roof.\label{i-live1}
    \item Roof Live Load: Roof live load is set as 20 psf surface load, different from the live load on other stories.\label{i-live2}
    \item Roof Dead Load: We assign 15 psf surface load for non-structure elements on roof panels.\label{i-dead2}
    \item Cladding Load: $20 \text{ psf} \times \text{$16$ ft (story height)}+90$ \nicefrac{lb}{ft} $= 410 $ \nicefrac{lb}{ft} line load is added to all boundary beams on each story for self-weights of cladding walls.\label{i-dead3}
    \item Modal Analysis: Modal analysis determines eigenvalues (eigenpulsations, eigenfrequencies, or eigenperiods), precision, eigenvectors, participation coefficients and participation masses for the problem of structural eigenvibrations. The number of modes is set to 30.
    \item Seismic X: Seismic loads are automatically computed by RSA given the building code. We consider seismic loads in two directions: X and Y. Settings of seismic loads are listed in Table \ref{tab:seismic_parameters}. Seismic X refers to the seismic loads in direction X.\label{i-eqx}
    \item Seismic Y: This is the seismic loads in direction Y.\label{i-eqy}
    \item Static Load Combination: Load combination linearly combines multiple load cases. Static load combination is defined as $1.2D + 1.6L +0.5L_{r}$, where $D$ is the sum of dead loads (\ref{i-self}+\ref{i-dead1}+\ref{i-dead2}+\ref{i-dead3}), $L$ is the live load (\ref{i-live1}),and $L_{r}$ is the roof live loads (\ref{i-live2}).
    \item Seismic Load Combination X: Complete quadratic combination (CQC) method is used for seismic load combination. This is defined as $0.9D + 1.0E_{x}$, where $E_{x}$ is the Seismic X load (\ref{i-eqx}).
    \item Seismic Load Combination Y: This is defined as $0.9D + 1.0E_{y}$, where $E_{y}$ is the Seismic Y load (\ref{i-eqy}).
\end{enumerate}

% Seismic Parameters
\begin{table}[t!]
\caption{Seismic Parameters}
\label{tab:seismic_parameters}
\vskip 0.15in
\begin{center}
\begin{small}  % \begin{sc}
\begin{tabular}{ll}
\toprule
Parameter Name & Value\\
\midrule
        Site Class & D\\
        $S_{1}$ (Acceleration parameter for 1-second period) & 0.6\\
        $S_{s}$ (Acceleration parameter for short periods.) & 1.8\\
        $Ie$ (Importance factor)& 0.0\\
        \hline
        Load to mass conversion for dead load & 1.0\\
        Load to mass conversion for live load & 0.1\\
        Load to mass conversion for roof live load & 0.25\\
\bottomrule
\end{tabular}  % \end{sc}
\end{small}
\end{center}
\vskip -0.2in
\end{table}

% number of nodes/edges
\begin{figure}[t!]
\centering     %%% not \center
\subfigure[Number of nodes distribution.] {\label{fig:num_of_node}\includegraphics[width=.75\columnwidth]{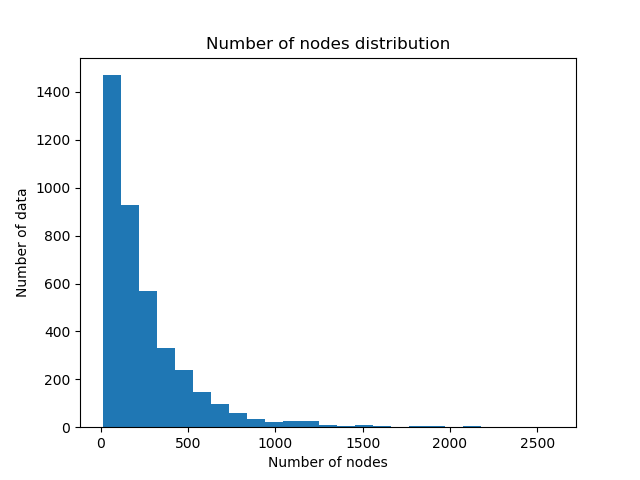}}
\subfigure[Number of edges distribution.] {\label{fig:num_of_edge}\includegraphics[width=.75\columnwidth]{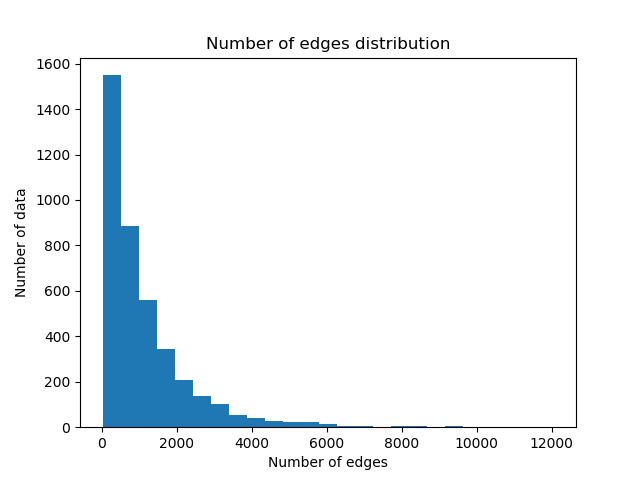}}
\caption{Statistics of 4000 collected structural graphs.}
\label{fig:datastatistics}
\end{figure}

% run/cal time
\begin{figure}[t!]
\centering     %%% not \center
\subfigure[Data collection time distribution.] {\label{fig:run_time}\includegraphics[width=.75\columnwidth]{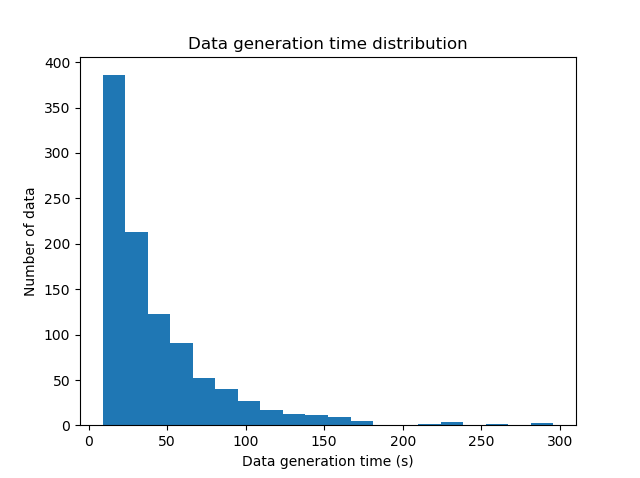}}
\subfigure[Calculation time distribution for solving structural simulations.] {\label{fig:cal_time}\includegraphics[width=.75\columnwidth]{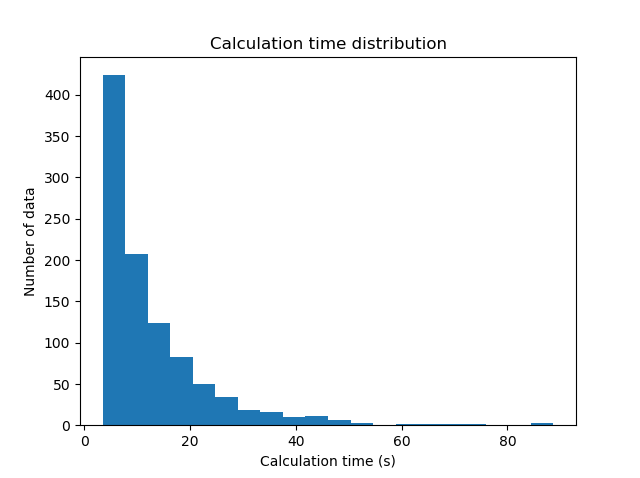}}
\caption{Statistics of 4000 collected structural graphs.}
\label{fig:datagenstatistics}
\end{figure}
% The average time to generate one datum is 43.91 seconds, out of which, 13.02 seconds are used to run structural simulation in RSA.  The average number of nodes is 252.79 while the average number of edges is 1132.27. Statistics histograms are attached in the supplementary materials.

\subsection{Saved Results}
After running the structural simulation, we save all drift ratios in direction X and Y for Seismic Load Combination X and Y load cases respectively. The components of the drift ratios perpendicular to the seismic load directions are relatively small compared to the drift ratio limit. The drift ratio distribution is normalized to $[-1,1]$. The open-source data set is available under \href{https://github.com/AutodeskAILab/LSDSE-Dataset}{https://github.com/AutodeskAILab/LSDSE-Dataset}.

\subsection{Statistics}
Figure \ref{fig:datastatistics} shows statistics of the collected 4000 building structural graphs. Figure \ref{fig:datagenstatistics} plots the two histograms: one for the times to generate one datum and the other for the calculation times spent on solving each structural simulation.

\section{Models Details}
\subsection{NeuralSim}
The input node feature of bar $i$ in a structural graph is denoted as $v_{i} \in \mathbb{R}^{19}$. A single layer perceptron (SLP) encoder first maps each node feature to embedding $ v^{0}_{i}\in \mathbb{R}^{512}$. 

\begin{equation}
v^{0}_{i} = SLP_{encoder}(v_{i})\label{eq:encoder}
\end{equation}

To perform message-passing in the propagation step, we first compute aggregated messages then update each node feature. The superscript $t$ denotes the propagation step and $Ne(i)$ denotes the neighbor set of node $i$.

\begin{gather}
  m_{i}^{t} = Mean\{SLP_{message}(v^{t}_{i}, v^{t}_{j}) | j\in Ne(i)\}
  \label{eq:neighbor_message}\\
  \begin{split}
      mp^{t}_{i}= Mean\{ SLP_{p\_message}([v_{i}^{t}, \frac{1}{d(i,j)+1} v_{j}^{t}])  , \\j= \mathop{\arg\max}_{l}\{d(l,i)|l\in A_{s}\} | \forall A_{s}, s=1\dots S\}
  \end{split} \label{eq:position_message}\\
  v^{t+1}_{i}= SLP_{update}(v^{t}_{i}, mp^{t}_{i},  m_{i}^{t}) \label{eq:update}
\end{gather}

Equation \ref{eq:neighbor_message} computes the aggregated message $m_{i}$ from neighbor nodes while Equation \ref{eq:position_message} computes the position-aware message $mp_{i}$ from the set of 512 anchor nodes $\{A_{s}| s=1\dots S\}$. $d(l,i)$ represents the geodesic distance between node $l$ and $i$. For more detail about position-aware message, we refer the readers to the \cite{you2019position}. Equation \ref{eq:update} updates each node embedding based on the current embedding and the messages. If position-aware message is not used, we drop $mp_{t}^{t}$ and change the input dimension of $SLP_{update}$ accordingly. In the end, we apply dropout function with 0.5 probability before the next propagation. In total, we run $T=5$ propagation steps.

Since NeuralSim is generating per-story output, we compute the story embedding $o_{k}$ by average pooling all the embeddings in story $k$.
\begin{gather}
o_{k} = AvgPool( \{ v^{T}_{i}) | i \in \textrm{Story } k  \} )\\
o_{k} \leftarrow SLP_{recursive}\{[o_{k}, o_{k+1}]\} \textrm{ for } k = K-1 \dots 1 \label{eq:recursive}
\end{gather}
Structured Decoder is processed using Equation \ref{eq:recursive}, where each story embedding is updated in the top-down order. In the end, the story embeddings are passed to two multi-layer perceptron (MLP)  decoders: one predicts the drift ratios $h_{k}\in \mathbb{R}^{2}$ and the other classifies if the ground-truth drift ratios exceed the drift ratio limit $lim=0.015$ or not. 
\begin{align}
    h_{k} &= MLP_{decoder}(o_{k})\label{eq:decoder1}\\
    c_{k} &= SigmoidMLP_{decoder}(o_{k})\label{eq:decoder2}
\end{align}

The multi-task loss is constructed by adding the L1 loss and the binary cross-entropy (BCE) loss.
\begin{equation}
    \textrm{Loss} = \frac{1}{K}\sum_{k=1}^{K}|h_{k}-\hat{h}_{k}| - w \times \text{BCE}(c_{k}, \hat{c}_{k})
\end{equation}
, where $\hat{h}_{k}$ is the ground-truth drift ratio , $\hat{c}_{k}$ is 1 if $\hat{h}_{k}>lim$, otherwise is 0, and $w=1$ is the weight balancing the two losses. NeuralSim is trained with 5 epochs, batch size 1, and learning rate $1\textrm{e-}4$ using the Adam optimizer.

% Algorithm \ref{alg:surrogate} shows the network structure of NeuralSim.
% % NeuralSim
% \begin{algorithm}[t!]
%   \caption{NeuralSim for Structural Simulation}
%   \label{alg:surrogate}
% \begin{algorithmic}
% %   \STATE {\bfseries Input:} node features $\{v_i\}$, dimension $19$
%     \STATE Encoder$([19,512])$
%     \FOR{$i=1$ {\bfseries to} $5$}
%     \STATE p\_Message$([1024, 512])$
%     \STATE Message$([1024, 512])$
%     \STATE Update$([1536, 512])$
%     \STATE Dropout
%     \ENDFOR
%     \STATE AvgPooling
%     \FOR{$i=1$ {\bfseries to} Max Story $K$}
%     \STATE recursive$([1024, 512])$
%     \ENDFOR
%     \STATE decoder$([512, 64, 16, 2], ReLU)$
%     \STATE sigmoid\_decoder$([512, 64, 16, 2], Sigmoid)$
% \end{algorithmic}
% \end{algorithm}

\subsection{NeuralSizer}
The inputs of the NeuralSizer are building skeleton geometries, which are represented as the same structural graphs except that the input node features $v_{i} \in \mathbb{R}^{10}$ now do not contain cross-sections. The encoder and propagation steps are the same as NeuralSim. Note that NeuralSizer does not compute nor use position-aware message by virtue of a faster training time. After 5 steps of propagation, the graph embedding $g$ is computed by MaxPooling all the node embeddings as below.
\begin{equation}
    g = MaxPooling({v_{i}^{T}|\forall i})
\end{equation}
Each node embedding together with the graph embedding is fed into an MLP decoder to generate one-hot vectors $y_{i}\in \mathbb{R}^{9}$ using hard Gumbel-Softmax function. The decoder has leaky ReLU function with negative slope 0.01 and dropout function in each layer.
\begin{equation}
    y_{i} = \text{GumbelSoftmax}(MLP_{decoder}(v_{i}^{T}, g))
\end{equation}

NeuralSizer is trained with batch size $5$ and learning rate $1\textrm{e-}4$ using Adam optimizer. 50,000 buildings are randomly sampled during training, and a fixed 500 data set is used for evaluation. The drop out probability is 0.5 and linearly decays to zero at the end of training. The loss function is given in the main paper.

% Algorithm \ref{alg:opt} shows the network structure of the NeuralSizer. 
% % NeuralSizer for Size Design
% \begin{algorithm}[tb]
%   \caption{NeuralSizer for Size Design}
%   \label{alg:opt}
% \begin{algorithmic}
%     \STATE Encoder$([10,512])$
%     \FOR{$i=1$ {\bfseries to} $4$}
%     \STATE Message$([1024, 512])$
%     \STATE Update$([1024, 512])$
%     \STATE ReLU
%     \STATE Dropout
%     \ENDFOR
%     \STATE MaxPooling
%     \STATE decoder$([1024, 64, 32, 10], ReLU)$
%     \STATE Gumbel-SoftMax
% \end{algorithmic}
% \end{algorithm}

\section{Dual Gradient Descent}
A general constrained optimization problem with an objective function $f(\theta)$ and an equality constraint $g(\theta)$ can be written as 
\begin{align}
    \mathop{\min}_{\theta}f(\theta) \qquad \text{s.t. } g(\theta)=0
\end{align}
Changing the constrained optimization to the dual problem, we get the Lagrangian:
\begin{equation}
    L(\theta, \lambda) = f(\theta) -\lambda g(\theta)
\end{equation}
, where $\lambda$ is the dual variable. Dual gradient descent alternates between optimizing the Lagrangian with respect to the primal variables to convergence, and then taking a gradient step on the dual variables. The necessity of optimizing the Lagrangian to convergence is optional under convexity. Both \cite{haarnoja2018soft} and our work found updating one gradient step still works in practice. As a result, the primal and dual variables are iteratively updated by the following equations.
\begin{align}
    \theta'&=\theta + \beta(\nabla_{\theta}f(\theta) -\lambda \nabla_{\theta}g(\theta))\\
    \lambda'&=\lambda + \gamma g(\theta)
\end{align}
where $\beta$ and $\gamma$ are learning rates. Inequality constraints can also be formulated similarly.

In this paper, our total loss is $w_{0}obj + w_{1}l_{dr} + w_{2}l_{var}+w_{3}l_{H}$. The initial weights and their learning rates are listed in Table \ref{tab:adaptive_weight_lr}.

% Dual Gradient Descent
\begin{table}[h!]   
\caption{Dual Gradient Descent Parameters}
\label{tab:adaptive_weight_lr}
% \vskip 0.in
\begin{center}
\begin{small}  % \begin{sc}
\begin{tabular}{lll}
\toprule
Loss & Initial Weight & Learning Rate \\
\midrule
Mass Objective & $w_{0}= 1,10$ &n/a\\
Drift Ratio Constraint & $w_{1}=1\textrm{e}3$ &$\gamma_{1}=1\textrm{e-}1$\\
Variety Constraint & $w_{2}=1.0$&$\gamma_{2}=5\textrm{e-}4$\\
Entropy Constraint &$w_{3}=1.0$ & $\gamma_{3}=1\textrm{e-}3$\\
\bottomrule
\end{tabular}  % \end{sc}
\end{small}
\end{center}
\vskip -0.1in
\end{table}

\section{Learning Curves for Neural Simulators}
\begin{figure}[h!]
  \centering
  \includegraphics[width=\columnwidth]{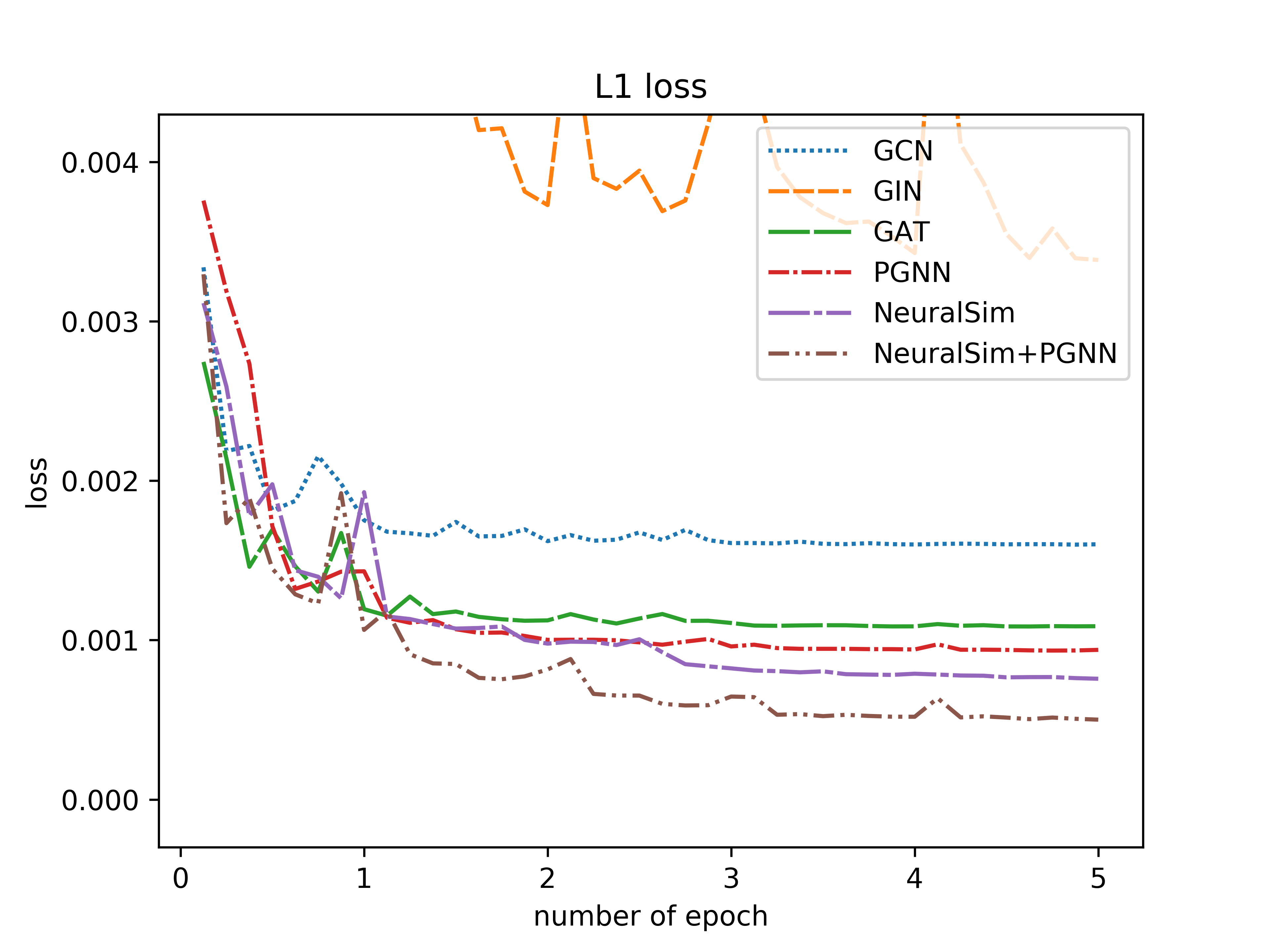}
  \caption{Learning curves of different models.}
  \label{fig:learning_curves}
\end{figure}

\begin{figure}[h!]
  \centering
  \includegraphics[width=\columnwidth]{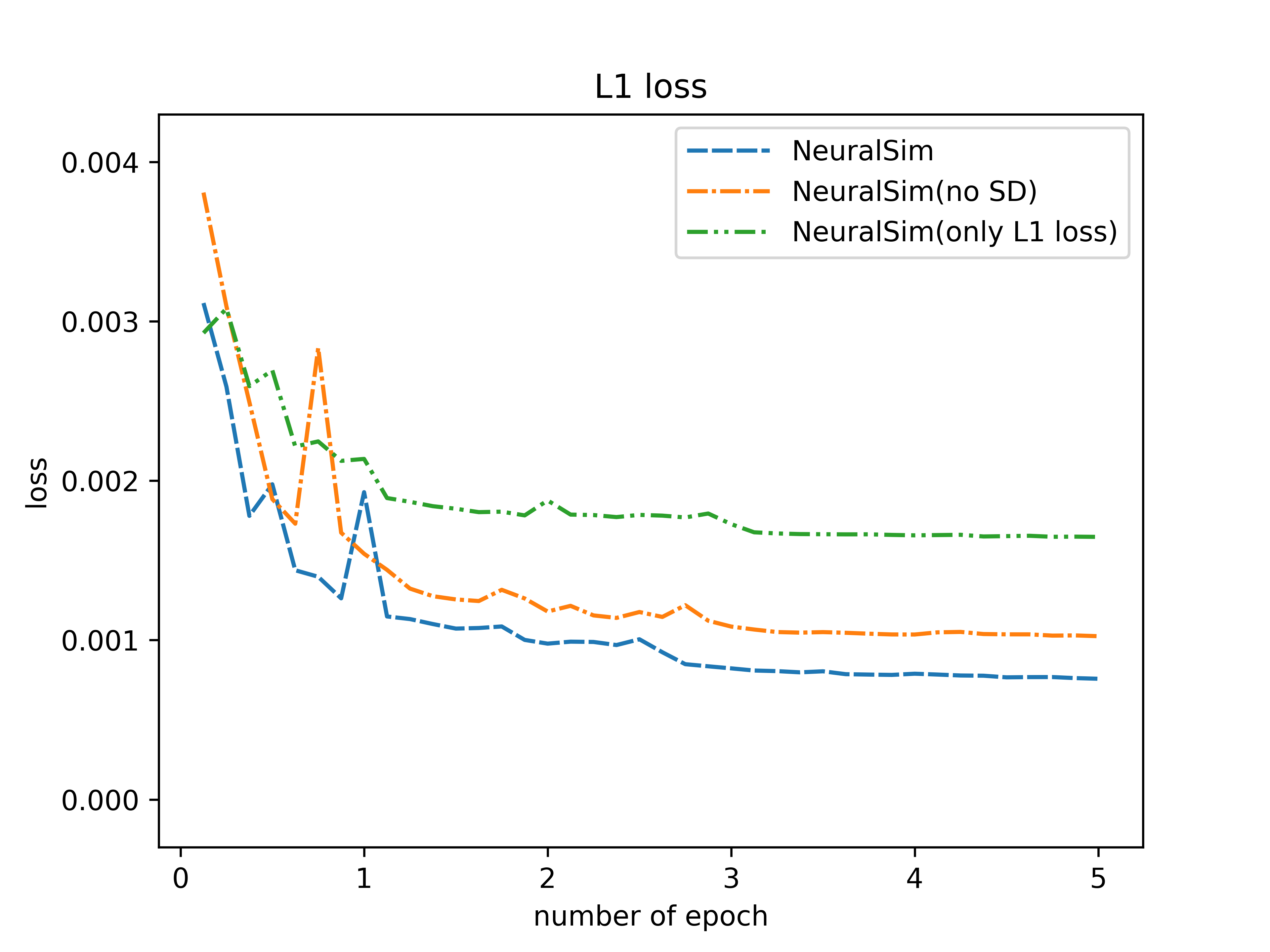}
  \caption{Learning curves of ablation studies.}
  \label{fig:learning_curves_ablation}
\end{figure}

Figure \ref{fig:learning_curves} and Figure \ref{fig:learning_curves_ablation} plot the learning curves of different models and ablation studies.

\section{User Study}
\begin{figure*}[t!]
  \centering
  \includegraphics[width=\textwidth]{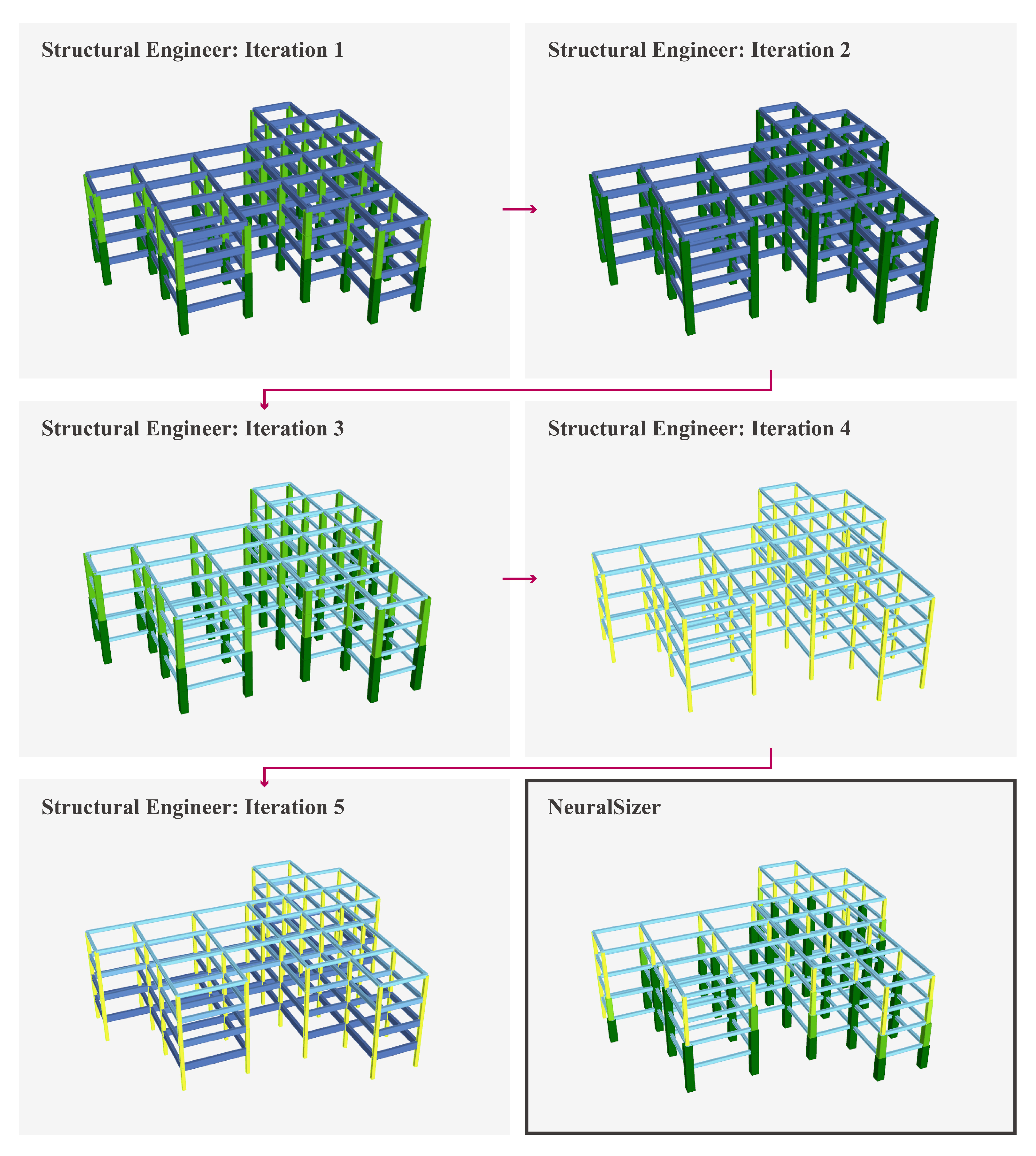}
  \caption{Visualization of NeuralSizer design and human designs.}
  \label{fig:userstudy}
\end{figure*}

We invite a structural engineer to work on a design in our user study and compare the human design with the design output from NeuralSizer. The structural engineer has 30 minutes to iterate on a 4-story building design following the manual design workflow. We also run NeuralSizer to create one design. All designs are evaluated with respect to the mass objective, the drift ratio constraint($<0.025$), and the variety constraint($<6$). In 30 minutes, the structural engineer is able to create five iterations, and the evaluation results are presented in Table \ref{tab:user_study}. The beam and column usages are sorted in the descending order of strength. Using stronger columns or beams lowers drift ratios, but leads to a larger total mass. The yellow-colored row highlights the most light-weighted designs that satisfy all constraints created by the structural engineer and NeuralSizer. The red cells indicate that the drift ratio exceeds the limit. The designs are visualized in Figure \ref{fig:userstudy}.

In the first two iterations, the structural engineer prioritizes the constraint satisfaction by choosing two designs using mostly the strongest columns and beams. The simulation results of the first two designs provide not only a baseline, but also a vague idea about the relation between the cross-section decisions and the drift ratio changes. The third design is the first attempt to optimize the mass objective and has a significantly smaller mass. However, when the structural engineer tries to further decrease the mass in the fourth and fifth iteration, both designs violate the  drift ratio constraint. Compared to the best human design in the third iteration, NeuralSizer outputs a design that has a lower mass while satisfying the constraints. Note that the 4-story building example is relatively simple. As the building becomes more massive (10 stories with more than 500 bars for example), the performance of human designs can degrade.

Potentially, human might still be able to create more optimal designs given more iterations. As a result, we claim that NeuralSizer can save the time and effort by providing a better initial design to structural engineers.  

\begin{sidewaystable}
% \centering
\caption{User Study Results}
\label{tab:user_study}
\begin{tabular}
{|m{3.0cm}|m{1.4cm}|m{3.5cm}|m{2.3cm}|m{0.8cm}|m{0.8cm}|m{0.8cm}|m{0.8cm}|m{0.8cm}|m{0.8cm}|m{0.8cm}|m{0.8cm}|}
\toprule
\multirow{2}{*}{Experiment} & 
\multirow{2}{*}{\parbox{1.2cm}{Mass\newline Tonnage}} &
\multirow{2}{*}{Beam Usage} & 
\multirow{2}{*}{Column Usage}&
\multicolumn{4}{|c|}{\parbox{3.6cm}{\centering Drift Ratios in Seismic X\newline(Limit $=0.025$)}}&
\multicolumn{4}{|c|}{\parbox{3.5cm}{\centering Drift Ratios in Seismic Y\newline(Limit $=0.025$)}}\\
\cline{5-12}
&&&& {\scriptsize Story 1}&{\scriptsize Story 2}&{\scriptsize Story 3}&{\scriptsize Story 4}& {\scriptsize Story 1}&{\scriptsize Story 2}&{\scriptsize Story 3}&{\scriptsize Story 4}\\
\hline
\multicolumn{12}{|c|}{Human}\\ 
\hline
Iteration 1 & 422.06 %10662.8
& \makecell[ccccccccc]{$188,0,0,0,0, 0, 0, 0, 0$}
& \makecell[ccccc]{$62, 62, 0, 0, 0$}
&0.0209&0.0174&0.015&0.0094
&0.0207&0.0173&0.0149&0.0093\\
Iteration 2 & 432.47 %11047.2
& \makecell[ccccccccc]{$188,0,0,0,0, 0, 0, 0, 0$}
& \makecell[ccccc]{$124, 0, 0, 0, 0$}
&0.0213&0.0178&0.0135&0.0084
&0.0211&0.0177&0.0134&0.0083\\
\rowcolor[HTML]{FFD700}
Iteration 3 & 279.41 %9710.8
& \makecell[ccccccccc]{$0,0,0,0,0, 0, 0, 0, 188$}
& \makecell[ccccc]{$62, 62, 0, 0, 0$}
&0.0208&0.0173&0.0149&0.0093
&0.0206&0.0172&0.0148&0.0093\\
Iteration 4 & 201.86 %6462.0
& \makecell[ccccccccc]{$0,0,0,0,0, 0, 0, 0, 188$}
& \makecell[ccccc]{$0, 0, 0, 0, 124$}
&\cellcolor[HTML]{F08080}0.0305&\cellcolor[HTML]{F08080}0.0253&0.0192&0.0119
&\cellcolor[HTML]{F08080}0.0302&\cellcolor[HTML]{F08080}0.0251&0.0191&0.0119\\
Iteration 5 & 267.95 %9604.3
& \makecell[ccccccccc]{$47,0,47,0,0,47, 0, 0, 47$}
& \makecell[ccccc]{$0, 0, 0, 0, 124$}
&\cellcolor[HTML]{F08080}0.0306&\cellcolor[HTML]{F08080}0.0253&0.0193&0.0119
&\cellcolor[HTML]{F08080}0.0303&\cellcolor[HTML]{F08080}0.0251&0.0191&0.0119\\
\hline
\multicolumn{12}{|c|}{NeuralSizer}\\ 
\hline
\rowcolor[HTML]{FFD700}
Objective Weight 10 & \cellcolor[HTML]{6cca2c}\textbf{257.47}%7078.3
& \makecell[ccccccccc]{$0, 0, 0, 0, 0, 0, 0, 0, 188$}
& \makecell[ccccc]{$ 71, 7, 2, 4, 40$}
&0.0196&0.0172&0.0149&0.0165
&0.0192&0.0198&0.0146&0.0161\\
\bottomrule
\end{tabular}
\end{sidewaystable}

\end{document}